%% file: _main.tex
\input{_constants}
\cameraready  

\pdfoutput=1
\documentclass[10pt,twocolumn,letterpaper]{article}
\input{cvpr_header}

\unless\ifarxiv \myexternaldocument{_supplementary} \fi

\usepackage{graphicx}
\usepackage{caption}
\usepackage{lipsum}

\begin{document}
\title{\paperTitle}
\author{\authorBlock}
\renewcommand{\thefootnote}{\fnsymbol{footnote}}

\twocolumn[{
\maketitle
\vspace{-0.8cm}
\centerline{\url{https://liuqi-creat.github.io/HOIGen.github.io}}
\begin{center}
    \captionsetup{type=figure}
    \includegraphics[width=.91\textwidth]{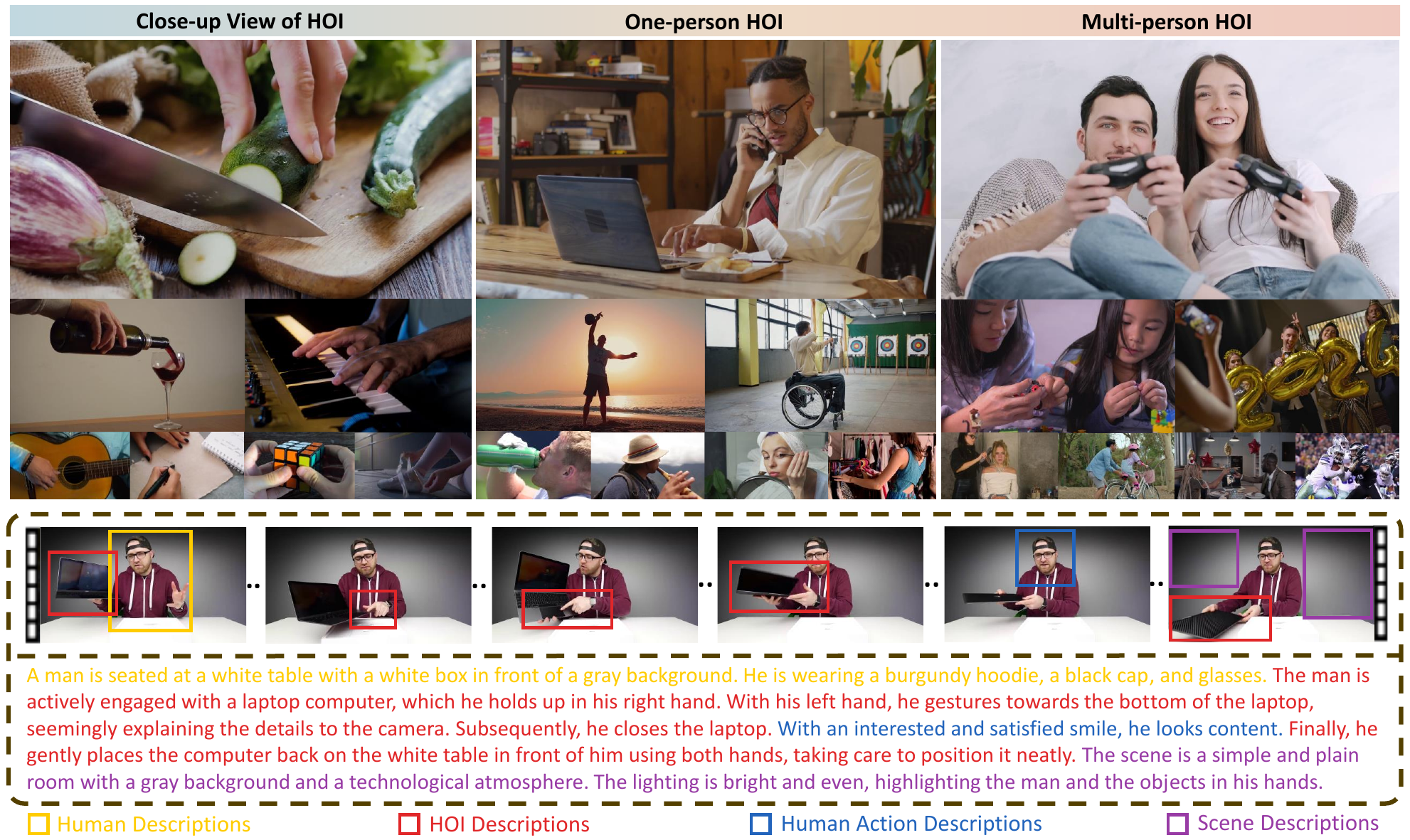}
    \captionof{figure}{Overview of HOIGen-1M. HOIGen-1M contains over one million video clips for HOI video generation with multiple types of HOI videos, diverse scenarios ($15,000+$ objects and $7,000+$ interaction types), and expressive captions.}
     \label{fig:dataset}
\end{center}
}]
\footnotetext[2]{Corresponding author.}

\input{00_abstract}
\input{01_intro}
\input{02_related}

\input{03_method}

\input{10_conclusion}

{\small
\bibliographystyle{ieeenat_fullname}
\bibliography{11_references}
}

\ifarxiv \clearpage \appendix \input{12_appendix} \fi

\end{document}


\title{\paperTitle}
\author{\authorBlock}
\maketitlesupplementary

\appendix
\input{12_appendix}

{\small
\bibliographystyle{ieeenat_fullname}
\bibliography{11_references}
}

%% file: _constants.tex
\def\paperID{4635}
\def\confName{CVPR}
\def\confYear{2025}

\def\paperTitle{HOIGen-1M: A Large-scale Dataset for Human-Object Interaction\\ Video Generation}

\def\authorBlock{
    Kun Liu\textsuperscript{1} \quad
    Qi Liu\textsuperscript{2} \quad
    Xinchen Liu\textsuperscript{1} \quad
    Jie Li\textsuperscript{1} \\
    Yongdong Zhang\textsuperscript{3} \quad
    Jiebo Luo\textsuperscript{4} \quad
    Xiaodong He\textsuperscript{1} \quad
    Wu Liu\textsuperscript{3} \footnotemark[2]\quad
    \\
    \textsuperscript{1}JD Explore Academy \quad 
    \textsuperscript{2}UCAS \quad   \textsuperscript{3}USTC \quad     \textsuperscript{4}University of Rochester 
    \\
}

\newif\ifreview 
\newif\ifarxiv 
\newif\ifcamera \newcommand{\cameraready}{\cameratrue}
\newif\ifrebuttal 

%% file: cvpr_header.tex
\ifreview \usepackage[review]{cvpr} \fi
\ifarxiv \usepackage[pagenumbers]{cvpr} \fi
\ifrebuttal \usepackage[rebuttal]{cvpr} \fi
\ifcamera \usepackage{cvpr} \fi

\input{_macros}  

\usepackage{xr-hyper}

\makeatletter
\newcommand*{\addFileDependency}[1]{
  \typeout{(#1)}
  \@addtofilelist{#1}
  \IfFileExists{#1}{}{\typeout{No file #1.}}
}

\makeatother
\newcommand*{\myexternaldocument}[1]{
    \externaldocument{#1}
    \addFileDependency{#1.tex}
    \addFileDependency{#1.aux}
}

\definecolor{cvprblue}{rgb}{0.21,0.49,0.74}
\usepackage[pagebackref,breaklinks,colorlinks,allcolors=cvprblue]{hyperref}
\usepackage[capitalize]{cleveref}
\crefname{section}{Sec.}{Secs.}
\crefname{table}{Table}{Tables}
\crefname{figure}{Fig.}{Figs.}

\ifarxiv \crefname{appendix}{App.}{Apps.}
\else \crefname{appendix}{Suppl.}{Suppls.} \fi

%% file: _macros.tex
\usepackage{graphicx}	
\usepackage{amsmath}	
\usepackage{amssymb}	
\usepackage{booktabs}
\usepackage{times}
\usepackage{microtype}
\usepackage{epsfig}
\usepackage{caption}
\usepackage{float}
\usepackage{placeins}
\usepackage{color, colortbl}
\usepackage{stfloats}
\usepackage{enumitem}
\usepackage{tabularx}
\usepackage{xstring}
\usepackage{multirow}
\usepackage{xspace}
\usepackage{url}
\usepackage{subcaption}
\usepackage{xcolor}
\usepackage[hang,flushmargin]{footmisc}
\usepackage{booktabs}
\usepackage{pifont}
\usepackage{paralist}

\ifcamera \usepackage[accsupp]{axessibility} \fi

% ...

\ifarxiv  \fi

\newcommand{\R}[1]{{%
    \textbf{%
        \ifstrequal{#1}{1}{\textcolor{red}{R#1}}{%
        \ifstrequal{#1}{2}{\textcolor{blue}{R#1}}{%
        \ifstrequal{#1}{3}{\textcolor{magenta}{R#1}}{%
        \ifstrequal{#1}{4}{\textcolor{teal}{R#1}}{%
                           \textcolor{cyan}{R#1}%
        }}}}%
    }%
}}

%% file: 00_abstract.tex
\begin{abstract}
Text-to-video (T2V) generation has made tremendous progress in generating complicated scenes based on texts. 
However, 
human-object interaction (HOI) often cannot be precisely generated by current T2V models due to the lack of large-scale videos with accurate captions for HOI. 
To address this issue, we introduce \textbf{HOIGen-1M}, the first large-scale dataset for \textbf{HOI} \textbf{Gen}eration, consisting of over one million high-quality videos collected from diverse sources. 
In particular, to guarantee the high quality of videos, we first design an efficient framework to automatically curate HOI videos using the powerful multimodal large language models (MLLMs), and then the videos are further cleaned by human annotators. 
Moreover, to obtain accurate textual captions for HOI videos, 
we design a novel video description method based on a Mixture-of-Multimodal-Experts (MoME) strategy that not only generates expressive captions but also eliminates the hallucination by individual MLLM. 
Furthermore, due to the lack of an evaluation framework for generated HOI videos, we propose two new metrics to assess the quality of generated videos in a coarse-to-fine manner.
Extensive experiments reveal that current T2V models struggle to generate high-quality HOI videos and confirm that our HOIGen-1M dataset is instrumental for improving HOI video generation.

\end{abstract}

%% file: 01_intro.tex
\section{Introduction}
\label{sec:intro}
Text-to-video (T2V) generation, which focuses on creating video sequences from descriptive text, is receiving increasing attention in computer vision. Due to the significant strides of large multi-modal models like Sora~\cite{sora}, T2V generation has recently been viewed as a promising route to building general simulators of the real world, offering vast potential for video creation, self-driving technology, and the creation of physical agents. 
Nevertheless, as a fundamental aspect of the physical world, human-object interaction (HOI) is challenging to generate accurately for current T2V models because large-scale datasets of videos with precise captions are absent. 
Therefore, this work offers a dataset, a captioning strategy, and automatic metrics for HOI video generation. 

\begin{figure}[t]
   \centering 
   \includegraphics[width = 1.0\linewidth]{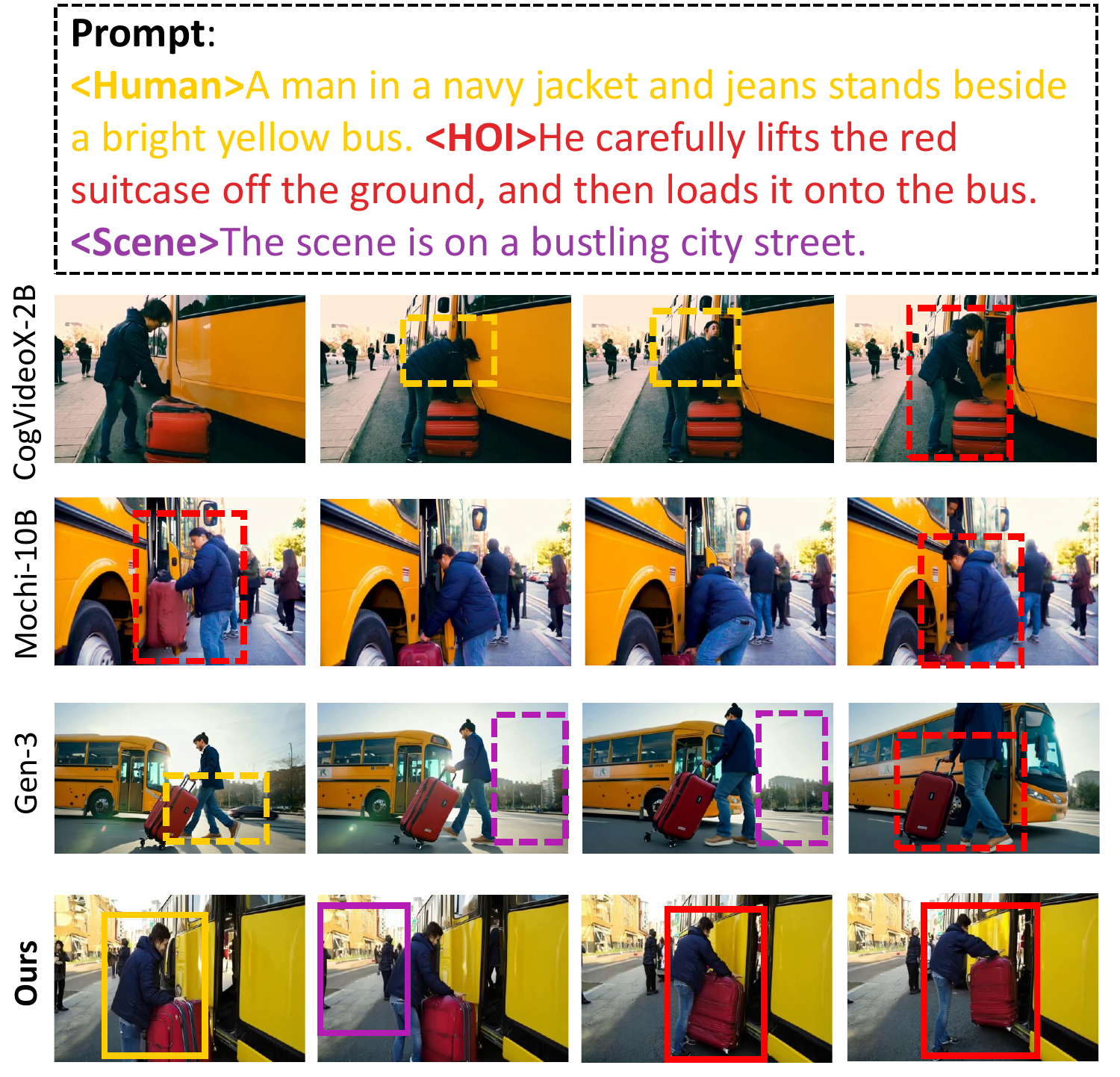}
    \caption{Video frames uniformly sampled from the video generated by several current T2V models. Most videos do not adequately follow the prompt of loading a suitcase onto a bus, showing that current T2V models struggle to generate videos that align with HOI. A dashed box means the content does not match the prompt, and a solid box indicates it does.}
   \label{fig:bingxiang}
   \vspace{-5mm}
\end{figure}

As shown in Figure~\ref{fig:bingxiang}, the current T2V models, even with over 10B parameters, still struggles to produce simple HOI videos.
Generally speaking, the videos' quality and the captions' accuracy are crucial to the performance of T2V models~\cite{yuan2024chronomagic,chen2024sharegpt4video}. 
However, current well-known video generation datasets, WebVid-10M~\cite{webvid} and Panda-70M~\cite{panda70m}, either have inferior quality or contain many videos without HOI. 
On the other hand, these video benchmarks for perceiving HOI are too small in scale to support the training of T2V models. 
Besides, existing captioning methods either produce descriptions that are too brief, restricting the in-depth video understanding, or they are not specifically designed HOI, resulting in the loss of many visual interaction details. Therefore, we aim to address the above issues by constructing a large-scale, high-quality training dataset for HOI video generation. 


The construction of a large-scale HOI dataset faces two main challenges. 
The first challenge is acquiring high-quality and extensive video data that includes HOI. 
This involves accurately sourcing videos that capture these interactions. 
The second challenge is obtaining high-quality captions that precisely describe the people, objects, and scenes involved. 
This requires accurate and detailed captions to convey the complexity of the interactions and settings depicted in the videos.
To address the aforementioned challenges, we build the first large-scale and high-quality dataset for HOI video generation named \textbf{HOIGen-1M}
(see Figure~\ref{fig:dataset}).
It exhibits three main features: 
1) Large scale: HOIGen-1M curates over 1M video clips and all videos contain manually verified HOI, which is sufficient for training T2V models. 
2) High quality: HOIGen-1M is strictly selected from the aspects of
mete attribute, aesthetics, temporal consistency, motion difference, and MLLM assessment. 
3) Expressive captions: the captions in HOIGen-1M are
precise because a Mixture-of-Multimodal-Experts (MoME) strategy is employed to detect and eliminate hallucinations via cross-verification among multiple MLLMs. 



To guarantee the above features of the dataset, 
we first design an efficient data pipeline, including multimodal large language model (MLLM) evaluation and human verification, to curate over 1 million videos showcasing HOI from a collection of 80 million videos. 
Next, we first propose a prompt engineering technique to drive MLLM to concentrate on describing human motion, object state changes, and HOI. More importantly, to mitigate hallucinations in video captioning by MLLMs, we design a Mixture-of-Multimodal-Experts (MoME) strategy to mutually verify the outputs among multiple MLLMs.  
Additionally, to accurately align human preference on the HOI video generation, we introduce two new automatic metrics, CoarseHOIScore and FineHOIScore, to assess the visual quality of interaction.
We hope this project can act as a crucial resource for HOI video generation. 

In summary, our work makes the following contributions:

\begin{compactitem}
	
    \item To our knowledge, we propose the first large-scale dataset for HOI video generation, comprising over one million high-quality and human-verified video clips. 
     \item To guarantee the quality of HOIGen-1M, we first design an efficient data pipeline to curate videos. Moreover, we develop to detect hallucinations and correct the errors in the captions via a Mixture-of-Multimodal-Experts (MoME) strategy.
          
     \item We introduce two new automatic metrics to evaluate the quality of HOI in the generated videos.
\end{compactitem}

Based on our dataset and evaluation framework, we compare ten T2V models with comprehensive analysis.
The experiments show that the performance of the open-source T2V model and commercial software in generating HOI videos is far from real-world applications.
Furthermore, all open-source models exhibit improved performance after fine-tuning on the constructed HOIGen-1M benchmark, which validates the value of this dataset for improving HOI video generation.
Specifically, an open-sourced model (Cog-VideoX-5B) fine-tuned HOIGen-1M achieves competitive performance to the most recently available commercial software Kling 1.5 in terms of  CoarseHOIScore. 


%% file: 02_related.tex
\begin{table*}[th]\footnotesize
\begin{center}
\newcommand{\tabincell}[2]{\begin{tabular}{@{}#1@{}}#2\end{tabular}}
\centering \caption{A Comparison of the existing datasets for the T2V task and Our HOIGen-1M.}
\label{tab:DatasetCompare}
\vspace{-2mm}
\begin{tabular}{ccccccc}
  \toprule
         Dataset   &\tabincell{c}{\#Videos}  &\tabincell{c}{Duration(Hours)}  &\tabincell{c}{Resolution}     &\tabincell{c}{Scenario}  &\tabincell{c}{Caption} &\tabincell{c}{Manually Filtering}\\
  \midrule       
      CelebV-HQ~\cite{zhu2022celebv}         &35K        &68    &512p     &Portrait         &N/A      &\ding{55}  \\
      CelebV-Text~\cite{yu2023celebv}         &7K        &279   &512p     &Face         &Short      &\ding{55}  \\
      ChronoMagic~\cite{yuan2024magictime}         &2K      &7    &720p     &Time-lapse   &Short &\ding{55}  \\
      ChronoMagic-Pro~\cite{yuan2024chronomagic}     &460K      &30K   &720p     &Time-lapse   &Short &\ding{55}  \\
      HowTo100M~\cite{miech2019howto100m}          &136M        &134.5K   &240p    &Instructional      &Short      &\ding{55}  \\
 \midrule
      MSR-VTT~\cite{xu2016msr}             &10K       &40    &240p    &General      &Short     &\ding{55}  \\  
      Panda-70M~\cite{panda70m}           &70M       &166.8K    &720p    &General      &Short    &\ding{55}  \\  
      WebVid-10M~\cite{webvid}          &10M       &52K    &360p     &General      &Short      &\ding{55}  \\  
      HD-VILA-100M~\cite{xue2022advancing}          &103M        &371.5K   &720p    &General      &Short      &\ding{55}  \\
      OpenVid-1M~\cite{nan2024openvid}          &1M        &2.0K   &{720p+}    &General      &Long      &\ding{55}  \\     
 \midrule
      HOIGen-1M                      &1M     &2.2K    &{720p+}    &HOI  &Long   &\ding{51} \\
  \bottomrule
\end{tabular}
\end{center}
\vspace{-7mm}
\end{table*} %

\section{Related Work}
\label{sec:related}

Our work touches two threads: text-to-video generation and HOI, which will be discussed respectively.

\textbf{Text-to-video Generation.}
With the development of the diffusion model and the Transformer architecture, 
Text-to-video generation has made tremendous progress in visualizing complicated scenes based on textual description~\cite{sora,Kling,opensora,opensoraplan,cogvideox}. 
For example, Sora~\cite{sora} is the first method to produce minute-long videos based on user prompts. 
Very recently, Kling 1.5~\cite{Kling} is an advanced video generation software that enables users to create high-quality videos through intuitive tools and customizable templates. 
Meanwhile, OpenSora~\cite{opensora}, OpenSoraPlan~\cite{opensoraplan}, and CogVideoX~\cite{cogvideox} leverage the million-scale video-text training samples to obtain the impressive results in the open-source community. 
While these models perform well in generating videos with general text, the quality of the generated HOI videos is far from satisfactory. 
Thus, it is necessary to construct a dedicated and large-scale dataset to generate HOI videos. 


Recent advancements in the T2V field have predominantly focused on data quality, leading to the emergence of numerous video and caption datasets within the research community~\cite{taichi,ChronoMagic,CelebV-Text,webvid,panda70m}. 
For instance, Taichi-HD~\cite{taichi} includes 2,668 videos of persons performing Taichi.
ChronaMagic~\cite{ChronoMagic} consists of 2,265 high-quality time-lapse videos with corresponding detailed captions. 
CelebV-Text~\cite{CelebV-Text} comprises 70,000 in-the-wild video clips with diverse visual content for facial video generation. 
In contrast, HOIGen-1M offers a million-level, high-quality collection for an unexplored domain. 
As for general video generation datasets, both WebVid-10M~\cite{webvid} and Panda-70M~\cite{panda70m} compile million-scale high-resolution and semantically consistent video samples. 
However, WebVid-10M~\cite{webvid} includes low-quality videos with watermarks and Panda-70M~\cite{panda70m} contains numerous static, flickering, low-clarity videos along with short captions, while our HOIGen-1M stands out as an precise T2V dataset tailored for HOI with excellent video quality and detailed captions. 
Table~\ref{tab:DatasetCompare} compares HOIGen-1M with existing datasets for the T2V task. 
\textbf{HOI Detection. }Numerous video-based datasets have been developed for HOI perception, such as HOI detection~\cite{cad, ActionGenome, MPHOI, MOMA, VHOI}.  
For instance, CAD-120~\cite{cad} is designed for HOI detection and comprises 120 RGB-D videos, limited to 10 high-level activities. 
MPHOI-72~\cite{MPHOI} focuses on multi-person activities, consisting of 72 videos captured from three distinct angles. 
The PVSG~\cite{PVSG} dataset contains 400 videos of both third-person and egocentric perspectives.

\textbf{HOI Reconstruction. }As the performance of HOI detection improves, HOI reconstruction datasets has also attracted extensive attention~\cite{GRAB, HOI4D, BEHAVE, HULC, COUCH, Compositional, xusemantic,diller2024cg}. 
For instance, the GRAB~\cite{GRAB} dataset comprises 1,334 full 3D shapes and pose sequences featuring 10 subjects engaging with 51 common objects. 
HOI4D~\cite{HOI4D} is a substantial 4D egocentric dataset comprising over 4,000 videos of nine participants interacting with 800 distinct object instances. 
BEHAVE~\cite{BEHAVE} is the first full-body HOI dataset to offer multi-view RGBD frames alongside corresponding 3D SMPL models, totaling 15.2K video frames.

Even though these above datasets provide benchmarks for HOI perception, they cannot be directly utilized for video generation. A training set of at least a million is typically necessary for T2V task, yet the current HOI benchmark comprises no more than tens of thousands of videos. Besides, existing HOI benchmarks lack high-quality videos and expressive textual descriptions, which are of great importance for the training of T2V models. 
In contrast, our HOIGen-1M dataset contains more than 1 million high-quality HOI videos with detailed captions.

%% file: 03_method.tex
\begin{figure*}[t]
   \centering 
   \includegraphics[width = 0.9\linewidth]{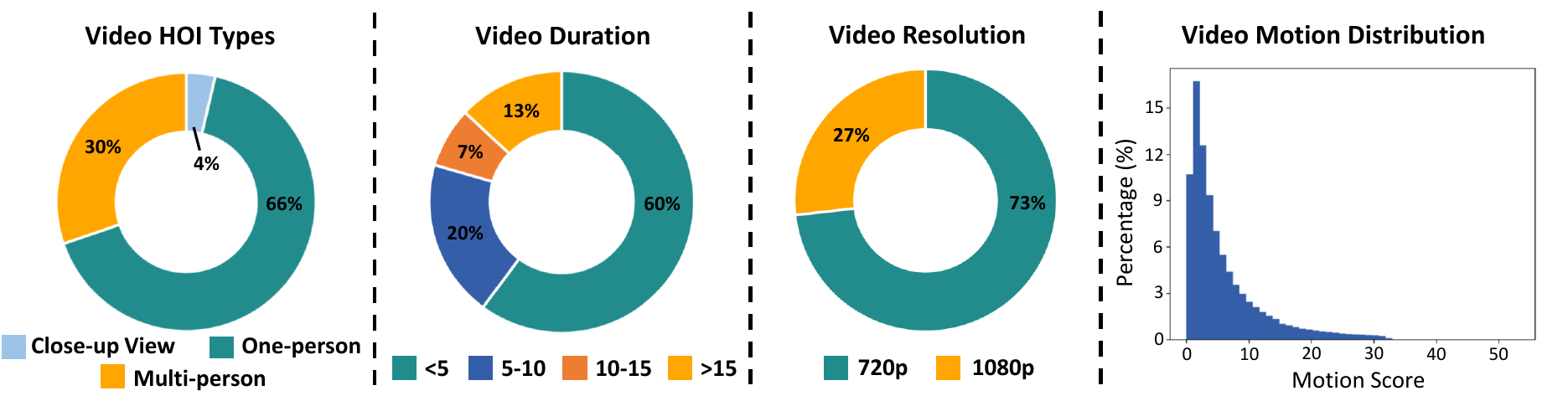}
    \caption{Statistics of video clips in HOIGen-1M. The dataset includes multiple types of HOI and spans a range of clip durations. All videos have a resolution of at least 720p and include significant motions.}
   \label{fig:dataset-only-video}
\end{figure*}

\section{The HOIGen-1M Dataset}
\label{sec:method}

\subsection{Video Curation and Processing}

\textbf{Data Collection.}
Initially, we select three public datasets designed for HOI perception: BEHAVE~\cite{BEHAVE}, InterCap~\cite{InterCap}, and HOI4D~\cite{HOI4D}, which have high resolution, and each clip showcases at least one HOI instance.
However, these datasets only contain 22K valid videos, which is far from sufficient to support the training of large video generation models.

To collect more HOI videos, we expand our dataset selection from HOI perception data to general videos.
We incorporate five large-scale datasets for T2V generation: Panda-70M~\cite{panda70m},  ViSR~\cite{liu2019social}, and Mixkit, Pixabay, and Pexels introduced in OpenSoraPlan~\cite{opensoraplan}. 
By this means, we obtain about eighty million raw videos.
However, these videos are of varying quality and most of them contain non-HOI content.
Therefore, to filter high-quality HOI videos from such a vast corpus, we adopt a data processing pipeline that is introduced as follows.

\textbf{Metadata.}
We initially analyze video metadata, including duration, resolution, and frames per second (FPS). 
Videos are filtered to retain those exceeding one second in duration, with a resolution $\ge$ 720p and a frame rate $\ge$ 20 FPS.

\textbf{Optical Character Recognition.}
The DBNet++~\cite{ocr} model is employed for optical character recognition to exclude video samples with over one text, which are considered as noises for the training of video generation models.

\textbf{Aesthetics Score.}
Video aesthetic scores are a key factor in determining the visual quality of videos. 
To ensure high-quality videos, we utilize the Laion Aesthetic Predictor~\footnote{https://github.com/christophschuhmann/improved-aesthetic-predictor} to assess the visual appeal of videos. 
Only those videos that achieve high aesthetic scores are retained.

\textbf{Motion Score.}
To further ensure video quality, we enhance video quality by computing smooth motion. 
Specifically, we employ UniMatch~\cite{flow} to calculate the optical flow scores of the videos. 
Videos with excessively high or low optical flow scores are excluded, while those with moderate scores are retained.



\textbf{LLM Evaluation.}
To efficiently identify videos containing HOIs, we first employ a multi-modal model, i.e., PLLaVA~\cite{PLLaVA}, to annotate each video with a detailed caption.
Then, we adopt a powerful LLM, i.e., Qwen2.5~\cite{qwen2.5} to judge whether there is an interaction in the caption using a predefined prompt.
After the above processing steps, we obtain about 1.5 million videos in total.


\textbf{Human Verification.}
To guarantee the high quality of the dataset, we employ seven annotators to manually check whether the remaining videos contain HOIs.
In particular, the annotators are asked to scan videos to remove those featuring inconspicuous HOI and minimal object visibility.
Finally, we obtain about 1.1 million videos from 80 million raw video clips.
The entire annotation process requires approximately eight weeks.

\textbf{Video Analysis.}
As illustrated in Table~\ref{tab:DatasetCompare}, our HOIGen-1M is a million-level, high-resolution, and manually verified video dataset for HOI generation.
Compared to previous datasets~\cite{zhu2022celebv,yu2023celebv,yuan2024magictime,yuan2024chronomagic} that are usually constructed for specific domains with limited videos, our HOIGen-1M is a large-scale dataset for an unexplored field. 
As shown in Figure~\ref{fig:dataset-only-video}, all videos feature resolutions of at least 720p and encompass three types of fine-grained interactions, including close-up view interaction, one-person interaction, and multiple-person interaction. 
In addition, compared to previous million-level datasets, WebVid-10M~\cite{webvid} includes low-grade videos with watermarks and Panda-70M~\cite{panda70m} comprises numerous static, flickering, or unclear videos. 
In contrast, our HOIGen-1M includes high-quality and clean videos without watermarks. 

\begin{figure*}[t]
   \centering 
   \includegraphics[width = 0.99\linewidth]{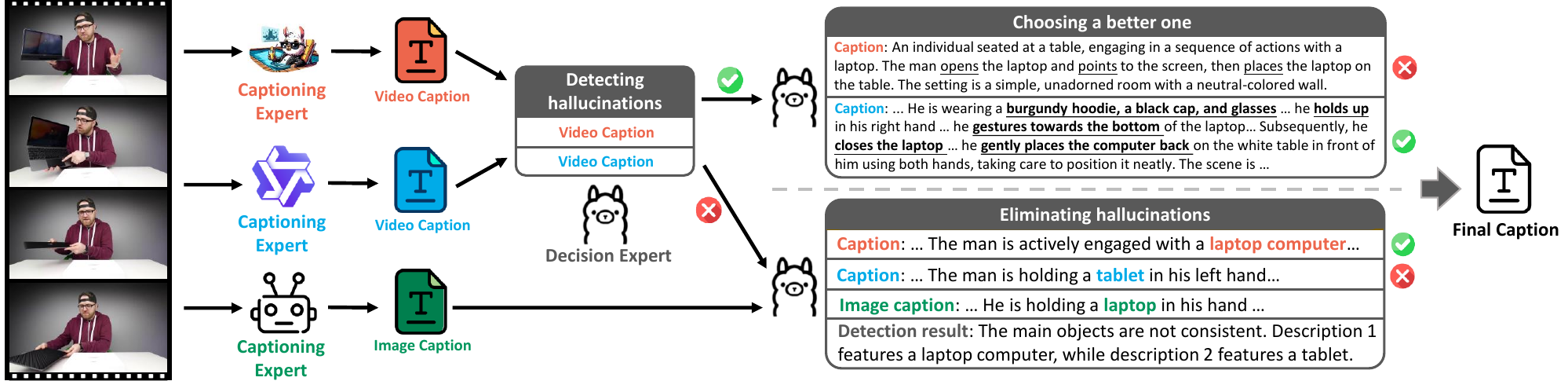}
    \caption{An illustration of the Mixture-of-Multimodal-Experts (MoME) strategy-based caption method.}
   \label{fig:caption}
\end{figure*}

\begin{figure*}[t]
   \centering 
   \includegraphics[width = \linewidth]{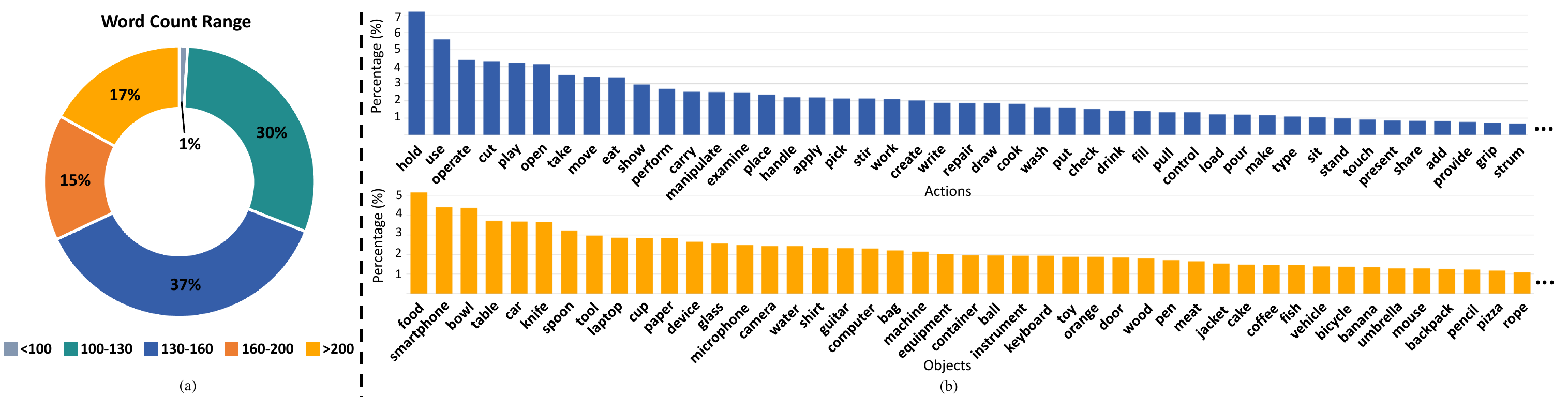}  
    \caption{Caption words statistics in HOIGen-1M. The distribution of word numbers shows the captions are high-quality and fine-grained, with an average length of 152 words. The distribution of actions and objects in the captions further demonstrates the diversity of the dataset. There are over 15,000 objects and over 7,000 interaction action types, making it possible to train a T2V model to simulate the real world. For clarity, we have only listed the categories with the highest frequency.}
   \label{fig:dataset-caption}
\end{figure*}

\subsection{Video Captioning with MoME}
The rich and detailed descriptions of videos are essential for training video generation models and the quality of generated videos.
To obtain the expressive caption, we adopt the powerful MLLMs to generate the textual description for the one million videos.
Specifically, we design a prompt engineering technique to adopt the MLLM for HOI description via a structured prompt, which encourages the model to focus on object attributes, human movements, and detailed temporal descriptions of interaction. 
However, the introduction of MLLMs may result in the hallucination phenomenon that the model generates plausible-sounding but factually incorrect or nonsensical texts. 
If we feed the video generation models with descriptions that are inconsistent with the visual content, the quality of the generated videos will be significantly degraded. 

To eliminate hallucinations generated by large models in video descriptions, we propose a Mixture-of-Multimodal-Experts Strategy (MoME) to detect hallucinations and then fuse the characteristics of different MLLM to correct them. 
As shown in Figure~\ref{fig:caption}, 
we first adopt two MLLMs, i.e., PLLaVa~\cite{PLLaVA} and Qwen2-VL~\cite{qwen2vl} as caption experts to generate multiple captions for one video. 
Then, we adopt Llama3.1~\cite{llama} as a decision expert to judge if both captions describe the same video, like the same object, interaction, and scene. 
A negative answer from the decision expert indicates a hallucination has been detected.
Upon detecting hallucination, we encourage the third expert ~\cite{liu2024visual} to focus on the areas where the first two experts have disagreements. 
At this point, the reason for the detected hallucination and the third caption are fed into the decision expert~\cite{llama} to regenerate the new caption, thus correcting the hallucination problem of the individual MLLM. 
If there is no hallucination, a decision expert is employed to evaluate which is superior, based on the richness of interactions, actions, and scene descriptions. 
In summary, MoME first adopts two captions and one decision expert to detect the hallucination. 
Then, an additional set of decision experts and caption experts is introduced to eliminate these hallucinations. 
Besides the textual description, we also annotate each video with HOI categories by extracting the core information from the caption. 
Thus, our dataset can also be utilized for HOI perception tasks. 

\textbf{Text Analysis. }
The captions in HOIGen-1M are longer and richer than those in WebVid-10M~\cite{webvid} and Panda-70M~\cite{panda70m}, where the average text lengths for WebVid-10M, Panda-70M, and HOIGen-1M are 12.0, 13.2, and 153.8, respectively.
Besides, as illustrated in Figure~\ref{fig:dataset-caption}, HOIGen-1M presents a wide variety of objects and interaction actions. 
There are over 15,000 objects and more than 7,000 interaction action types, which significantly exceeds the categories of HOI datasets. Thus, HOIGen-1M is a natural testbed for open-vocabulary HOI detection. 
Compared with previous video-text datasets, the captions in HOIGen-1M are precise because the MoME is employed to eliminate hallucinations via cross-verifications among multiple MLLM.  
Moreover, the interaction diversity within HOIGen-1M makes it possible to train a T2V model to simulate the real world.

\section{Experiments}
\label{sec:experiments}
In this section, we introduce the designed evaluation metrics and evaluate ten popular T2V models for HOI video generation. Moreover, we conduct an ablation study to highlight the effectiveness of HOIGen-1M and the proposed caption method. 
For implementation details, please refer to the \textbf{Supplementary File}. 

\subsection{Evaluation Protocol and Metrics} 
To make the experiment more convincing, we first employ the widely used evaluation metric VBench~\cite{huang2024vbench}. Because some of the 16 evaluation dimensions, such as color, are related to the prompts, 
we can adopt seven available metrics from VBench to assess the quality of the generated videos. 
Although the existing evaluation benchmarks are comprehensive, there are no specific metrics to assess HOI. 
Therefore, we first build a prompt set and then propose two evaluation metrics to assess the quality of generated interactive video clips in a coarse-to-fine manner.  

\begin{table*}[th]\footnotesize
\begin{center}
\newcommand{\tabincell}[2]{\begin{tabular}{@{\hskip-0.8pt}@{\hskip--0.8pt}@{\hskip-0.8pt}@{\hskip-0.8pt}@{\hskip-0.8pt}@{\hskip-0.8pt}#1@{\hskip-0.8pt}@{\hskip-0.8pt}}#2\end{tabular}}
\centering \caption{Evaluation results with proposed HOIScores and VBench in HOI video generation.}
\label{tab:evaluation9models}
\vspace{-2mm}
\begin{tabular}{cccccccc|cc}
  \toprule
         Method   &\tabincell{c}{Subject\\Consistency}  &\tabincell{c}{Background\\Consistency}  &\tabincell{c}{Motion\\Smoothness}  &\tabincell{c}{Aesthetic\\Quality} &\tabincell{c}{Imaging\\Quality} &\tabincell{c}{Temporal \\ Flickering} &\tabincell{c}{Overall\\Consistency}  &\tabincell{c}{Coarse\\HOIScore} &\tabincell{c}{Fine\\HOIScore}\\
  \midrule  
      OpenSoraPlan~\cite{opensoraplan}           &\textbf{97.06\%}       &\textbf{96.83\%}    &\textbf{99.56\%}       &55.43\%     &59.62\%      &\textbf{99.09\%}  &14.12\%  &8.80\%  & 91.54\%\\
      Mochi-10B~\cite{genmo2024mochi}	&87.27\%	&92.68\%	&98.90\%	&54.22\%	&54.31\%	&97.94\%	&19.83\%	&28.78\%	&\textbf{96.15\%}\\
      CogVideoX-2B~\cite{cogvideox}           &91.31\%       &95.17\%    &98.08\%      &53.48\%    & 60.94\%     &  96.13\% &20.29\% & 31.34\%   &92.06\%\\  
      OpenSora~\cite{opensora}             &94.33\%       &96.05\%    &99.05\%        &53.79\%     &\textbf{64.84\%} &98.52\%  &18.94\%  & 31.86\%  & 91.86\%\\  
      CogVideoX-5B~\cite{cogvideox}          &90.29\%        &94.96\%    &97.68\%         &\textbf{54.84\%}     &62.27\%      &95.41\% &\textbf{20.54\%}  & \textbf{32.84\%}  & 94.25\%\\  
 \midrule
     Gen-3~\cite{Gen3}           &90.37\%      & 95.31\%   &99.24\%        &59.22\%     &64.56\%    &97.95\%   &17.48\%   & 23.98\% &96.79\%\\  
     Pika~\cite{Pika}           &96.10\%      & \textbf{97.22\%}   &\textbf{99.51\%}        &60.55\%    &66.77\%      &\textbf{98.86\%}  &17.29\%  & 29.37\%  &97.44\%\\  
     Dreamina~\cite{Dreamina}             &\textbf{96.46\%}      &97.08\%    &98.85\%       &\textbf{66.07\%}     &67.99\%  &97.46\%   & 18.02\%   & 36.36\%  &94.18\%\\  
     Hailuo~\cite{Hailuo}          &92.61\%       &95.45\%    &99.16\%         &58.49\%     & \textbf{70.07\%}     &97.73\% & \textbf{19.55\%}  & 39.56\% &97.22\%\\            
     Kling 1.5~\cite{Kling}             &92.55\%       &95.42\%   &99.24\%       & 61.09\%    & 69.28\% & 96.70\% & 18.74\%  & \textbf{42.72\%} & \textbf{97.76\%}\\  
  \bottomrule
\end{tabular}
\end{center}
\vspace{-5mm}
\end{table*}

\textbf{Prompt Suite. }The first step in evaluating T2V models is to construct a comprehensive and diverse prompt suite. 
Inspired by these previous works~\cite{huang2024vbench,meng2024towards}, we select the object and interaction types from a large-scale HOI detection dataset~\cite{chao2015hico}. 
Then, we manually filter out rare and unreasonable interactions and get a total of 306 prompts, including interaction with musical instruments, transportation, kitchen tools, etc. 
Only interactive-type phrases are not sufficient for the model to generate promising videos. 
Thus, LLM is utilized to extend prompts longer and more detailed without changing their original meaning. 

\textbf{The Proposed CoarseHOIScore. }
An ideal interaction should at least involve people, objects, and corresponding actions.
Therefore, we adopt a HOI detector $d(\cdot)$~\cite{detector} to predict possible HOI triplets in the generated videos. 
Specifically, for each video, the CoarseHOIScore is calculated as:
\begin{equation}
    S_{coarse} = \frac{1}{N} \sum_{i=1}^{N} \left\{
      \begin{array}{ll}
        1 & \text{if } d(f_i) > \theta, \\
        0 & \text{otherwise},
      \end{array}
    \right.
\end{equation}
where $S_{coarse}$ is the CoarseHOIScore of one video, $N$ is the number of uniformly adopted video frames, $f_i$ denotes the $i_{th}$ frame, and $\theta$ is the threshold for the confidence score.
We average the CoarseHOIScore for all the videos generated by one model as the final score of this model.

 
\textbf{The Proposed FineHOIScore. }
Although the above operation can roughly determine interactions, it does not capture the details well, such as the presence of contact between the human and the object, and the stability of the interaction action.
To solve this problem, we propose a fine-grained, pixel-level approach to evaluate the generated videos.
Specifically, for each video, the FineHOIScore is calculated as:
\begin{equation}
    S_{Fine} = 1-\frac{1}{T-1} \sum_{i=2}^{T}(|\langle M_i, K_i\rangle - \langle M_{i-1}, K_{i-1}\rangle|),
\end{equation}
where $S_{Fine}$ is the FineHOIScore of one video, $T$ is the number of video frames, $M_i$ and $K_i$ denote the object mask and human keypoints in the $i^{th}$ frame, and $\langle \cdot \rangle$ represents the pixel distance that satisfies the requirement of the contact between people and objects. 
We average the FineHOIScore for all the videos generated by one model as the final score of this model.

\begin{table*}[th]\footnotesize
\begin{center}
\newcommand{\tabincell}[2]{\begin{tabular}{@{\hskip-0.8pt}@{\hskip-0.8pt}@{\hskip-0.8pt}@{\hskip-0.8pt}@{\hskip-0.8pt}@{\hskip-0.8pt}#1@{\hskip-0.8pt}@{\hskip-0.8pt}}#2\end{tabular}}
\centering \caption{The effect of the proposed captioning method.}
\label{tab:caption-effect}
\vspace{-2mm}
\begin{tabular}{cccccccc|cc}
  \toprule
         Method   &\tabincell{c}{Subject\\Consistency}  &\tabincell{c}{Background\\Consistency}  &\tabincell{c}{Motion\\Smoothness}     &\tabincell{c}{Aesthetic\\Quality} &\tabincell{c}{Imaging\\Quality} &\tabincell{c}{Temporal \\ Flickering} &\tabincell{c}{Overall\\Consistency}  &\tabincell{c}{Coarse\\HOIScore} &\tabincell{c}{Fine\\HOIScore}\\    
    \midrule     
      Fine-tuned with PLLaVA~\cite{PLLaVA}           &90.53\%	&94.29\%	&96.51\%	&52.91\%	&62.63\%	&95.24\%	&\textbf{19.93\%}  &33.17\%  &92.78\%\\    
      Fine-tuned with our Caption            &\textbf{94.98\%}	&\textbf{96.63\%}	&\textbf{98.16\%}	&\textbf{54.90\%}	&\textbf{65.85\%}	&\textbf{97.94\%}	&19.55\%	&\textbf{39.13\%}  &\textbf{93.82\%}\\  
  \bottomrule
\end{tabular}
\end{center}
\vspace{-3mm}
\end{table*}

\begin{table*}[th]\footnotesize
\begin{center}
\newcommand{\tabincell}[2]{\begin{tabular}{@{\hskip-0.9pt}@{\hskip--0.8pt}@{\hskip-0.8pt}@{\hskip-0.8pt}@{\hskip-0.9pt}@{\hskip-0.8pt}#1@{\hskip-0.8pt}@{\hskip-0.8pt}}#2\end{tabular}}
\centering \caption{The effect of fine-tuning using HOIGen-1M.}
\label{tab:finetunedDataset}
\vspace{-2mm}
\begin{tabular}{cccccccc|cc}
  \toprule
         Method   &\tabincell{c}{Subject\\Consistency}  &\tabincell{c}{Background\\Consistency}  &\tabincell{c}{Motion\\Smoothness}    &\tabincell{c}{Aesthetic\\Quality} &\tabincell{c}{Imaging\\Quality} &\tabincell{c}{Temporal \\ Flickering} &\tabincell{c}{Overall\\Consistency}  &\tabincell{c}{Coarse\\HOIScore} &\tabincell{c}{Fine\\HOIScore}\\    
    \midrule     
      OpenSora~\cite{opensora}             &94.33\%       &96.05\%    &99.05\%      &53.79\%     &\textbf{64.84\%} &98.52\%  &18.94\%  & 31.86\% &91.86\%\\ 
     Fine-tuned           &\textbf{96.94\%}	&\textbf{97.21\%}	&\textbf{99.28\%}		&\textbf{54.29\%}	&64.74\%	&\textbf{99.13\%}	&\textbf{19.36\%}	&\textbf{35.38\%} &\textbf{94.91\%}\\  
     \midrule     
      CogVideoX-2B~\cite{cogvideox}               &91.31\%    &95.17\%    &98.08\%    &53.48\%    & 60.94\%   &96.13\%    &\textbf{20.29\%}    & 31.34\% &92.06\% \\       
      Fine-tuned            &\textbf{94.98\%}	&\textbf{96.63\%}	&\textbf{98.16\%}	&\textbf{54.90\%}	&\textbf{65.85\%}	&\textbf{97.94\%}	&19.55\%	         &\textbf{39.13\%}  &\textbf{93.82\%}\\  
     \midrule 
      CogVideoX-5B~\cite{cogvideox} &90.29\%        &94.96\%    &97.68\%       &54.84\%     &62.27\%      &95.41\% &\textbf{20.54\%}  & 32.84\% &94.25\%\\
      Fine-tuned  &\textbf{94.70\%}	&\textbf{96.42\%}	&\textbf{98.29\%}	&\textbf{56.14\%}	&\textbf{66.46\%}	&\textbf{98.01\%}	&19.88\%	&\textbf{44.04\%} &\textbf{96.04\%}\\
  \bottomrule
\end{tabular}
\end{center}
\vspace{-3mm}
\end{table*}

\subsection{Main Results of HOI Video Generation}
Despite the recent significant attention on T2V benchmarks, systematic evaluation of these models on HOI video generation is still lacking. 
To solve this problem, we evaluate five popular commercial software Kling 1.5~\cite{Kling}, Pika~\cite{Pika}, Hailuo~\cite{Hailuo}, Dreamina~\cite{Dreamina}, and Gen-3~\cite{Gen3}, as well as five representative open-source methods including OpenSora~\cite{opensora}, OpenSoraPlan~\cite{opensoraplan}, CogVideoX-2B~\cite{cogvideox}, CogVideoX-5B~\cite{cogvideox}, and Mochi-10B~\cite{genmo2024mochi}.

As illustrated in Table~\ref{tab:evaluation9models}, even the recent advanced commercial software, Kling 1.5~\cite{Kling}, only attains a score of 42.72\% in terms of CoarseHOIScore. 
This indicates that even when trained on extremely large-scale open-domain datasets, existing T2V models still cannot produce clips that perfectly align with HOI. 
Consistent with the above observation, Michi-10B~\cite{genmo2024mochi}, one of the largest current text-to-video models with its 10 billion parameters, does not achieve an impressive HOI score for video generation. 
This indirectly reflects the necessity of constructing a dedicated and large-scale dataset to generate HOI videos.

In addition, we identify the following important observations: 1) While VBench provides a comprehensive evaluation standard, it does not accurately reflect the quality of the generated interactions. For instance, the score of motion smoothness over all models are more than 98\% but the generated interactions in videos are clearly unreasonable. 
It indicates the necessity of designing a standard specifically for evaluating the quality of interaction generation; 
2) Among open-source approaches, CogVideoX-5B~\cite{cogvideox} and Mochi-10B~\cite{genmo2024mochi} perform comparatively well, 
both exceeding the performance level of OpenSoraPlan~\cite{opensoraplan} and OpenSora~\cite{opensora} in terms of HOIScore. 
In relation to the scaling law of model parameters, the CogVideoX-5B model exhibits superior performance compared to the CogVideoX-2B model; 
3) All commercial software performs well in the VBench evaluation. However, there is significant variation in the HOI evaluation, with the best software achieving a score of 42.72\% and the worst only 23.98\%; 
4) Overall, under evaluation metrics VBench and HOIScore, commercial software tends to perform better than open-source software, possibly because commercial companies have more computational resources and easier access to large-scale video data for training models. 
For human evaluation on massive generated videos, please refer to the \textbf{Supplementary File}.

\subsection{Ablation Study} 
We conduct a thorough effectiveness analysis of the proposed captioning method to demonstrate its ability to generate high-quality textual descriptions, which in turn contributes to generating videos that align with HOI. 
We also demonstrate the advantage of our HOIGen-1M by fine-tuning T2V models on this dataset. 
We observe a significant improvement in the quality of HOI generation. 
This improvement highlights the dataset's comprehensive coverage and its ability to enhance model performance by providing rich and diverse examples. 
The combination of our captioning method and the HOIGen-1M dataset sets a new benchmark in the field, offering a powerful tool for future research and applications in HOI modeling. 

\noindent \textbf{The effectiveness of the proposed captioning method. }
We adopt the recently proposed PLLaVA~\cite{PLLaVA} as the baseline model due to its expressive performance on video captioning. 
Based on this model, we employ the proposed MoME to detect and eliminate hallucinations of individual MLLM, yielding a precise textual description. 
We train the T2V model separately using PLLaVA and our caption, and the evaluation results of the generated videos are listed in Table~\ref{tab:caption-effect}. We can observe that the T2V model trained with our caption achieves significant improvement in both VBench and HOI evaluation. 
Notably, our caption increases the CoarseHOIScore from 33.17\% to 39.13\%.

\noindent\textbf{The effectiveness of HOIGen-1M dataset. }
The quality of the generated interactive videos significantly improves after fine-tuning the model on HOIGen-1M. 
Table~\ref{tab:finetunedDataset} outlines the performance of the original model compared to its results after fine-tuning. 
Specifically, we first evaluate three models pre-trained on large-scale but general video samples. 
Then, we fine-tune all models on the HOIGen-1M using the LORA mechanism. 
A significant increase can be observed after fine-tuning in the HOI scores of all three models, 
which directly demonstrates the effectiveness of our dataset in generating HOI videos. 
In addition, we hope that our dataset can serve as a benchmark for interactive video generation.

\begin{figure*}[t]
   \centering 
   \includegraphics[width = 1.0\linewidth]{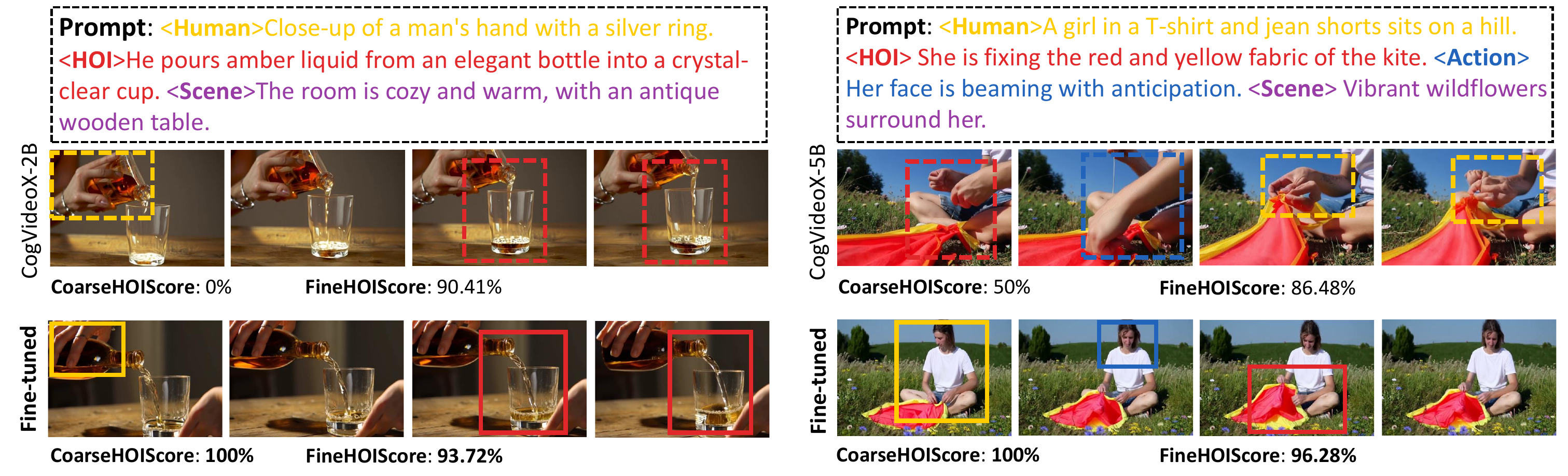}
    \caption{Visual comparison between CogVideoX and its fine-tuned version on HOIGen-1M. Please zoom in for more details.}
   \label{fig:finetuned}
\vspace{-5mm}
\end{figure*}

Besides quantitative results, we would like to obtain further insight into the learned T2V models. 
In this sense, we provide visual comparisons between CogVideoX and its fine-tuned version on HOIGen-1M in Figure~\ref{fig:finetuned}. 
According to the left example, we can see that the hands in the second row appear more natural, the cup is more aesthetically pleasing, and the water increases as the pouring progresses. 
The right example also demonstrates that our model generates clearer hands and more reasonable interaction with kites after fine-tuning. 
The CoarseHOIScores of the original CogVideoX model in both examples are low (0\% and 50\%) because the generated hands are too terrible to be detected. 
Both HOI scores demonstrate substantial improvement after fine-tuning. 
The above observations further directly demonstrate the effectiveness of our dataset.

\noindent\textbf{Best vs. Worst Generated Videos. }
Based on the HOIScore, we select the high-scoring interaction types and the low-scoring interaction types from these generated videos. 
For example, the interaction carrying a chair achieves a score of 87.5\%, which far exceeds the average score of all generated videos. 
In contrast, the interaction brushing receives a very low score, 
possibly because the detailed body movements required by this action are difficult to model. 
We can observe that simple actions and interactions with rigid objects generally achieve better generation results. 
Adopting prior knowledge (e.g., human keypoints) as a prerequisite for generation can help the model better understand and capture complex body movements, thereby improving the accuracy and reliability of the generated content. 
More visual results can be found in the \textbf{Supplementary File}. 

\noindent\textbf{Out-of-Distribution vs. In-Distribution Video Generation. }
Although we aim to build a dataset that encompasses as many HOI as possible to simulate the physical world, there are still some interactions that rarely appear in HOIGen-1M. This provides us with an opportunity to test the emergent capabilities of the T2V model. 
Specifically, we select 20 prompts from all prompts that have not appeared in the captions of HOIGen-1M as the out-of-distribution samples. 
Additionally, we choose 20 prompts that appear with the highest frequency in the HOIGen-1M as in distribution samples. 
We then calculate the HOIScores of the videos generated from these two groups of prompts. 
The close scores (63.20\% vs. 61.20\%) demonstrate that whether a prompt has been seen in the training set does not significantly influence the generation results. 
This might indicate that the model is beginning to demonstrate emergent capabilities, as it can generate interactions it has not encountered before. 

%% file: 10_conclusion.tex
\section{Discussion} 

While this work offers a high-quality dataset and an evaluation framework for HOI video generation, it still has two limitations. 
First, million-level videos may also be insufficient to support the training of T2V models as model sizes continue to scale up. 
Hence, we will further extend HOIGen-1M by collecting more high-quality videos and annotating more precise captions.
Second, although two new evaluation metrics, i.e., CoarseHOIScore and FineHOIScore, are introduced to assess the quality of generated HOI videos, there is still a disparity between these metrics and human preferences. 
In future work, we will design more comprehensive metrics aligned with human judgment to evaluate T2V models.


\section{Conclusion}
\label{sec:conclusion}
This paper proposes HOIGen-1M, a large-scale and high-quality dataset for HOI generation. 
To ensure video quality, we develop an efficient data selection pipeline to guarantee that each video clip contains a single clear HOI through MLLM evaluation and human verification. 
Besides, we meticulously design a Mixture-of-Multimodal-Experts (MoME) strategy to describe the HOI in detail, which can detect and then correct the hallucinations in captions generated by individual MLLM. 
Due to the lack of evaluation metrics for HOI video generation, 
we propose a novel assessment framework to assess the generated videos in coarse and fine manners.  
The experimental results and in-depth analysis demonstrate
the effectiveness of HOIGen-1M and the proposed caption method. 
With its data, annotations, and evaluation metrics, 
we believe HOIGen-1M can fuel future studies for HOI video generation.

\section{Acknowledge}
\label{sec:acknowledge}
This work is supported by National Key Research and Development Program of China (NO. 2024YFE0203200),  the National Nature Science Foundation of China (NO. U24A20329)

%% file: 12_appendix.tex
\section{Appendix Section}
\label{sec:appendix_section}
Supplementary material goes here.

\section{Appendix Section}
\label{sec:appendix_section}
Supplementary material goes here 2.